\documentclass[journal,12pt,onecolumn,draftclsnofoot,]{IEEEtran}
\usepackage{amsmath,amssymb,amsfonts}
\usepackage{algorithmic}
\usepackage{array}
\usepackage[caption=false,font=normalsize,labelfont=sf,textfont=sf]{subfig}
\usepackage{textcomp}
\usepackage{stfloats}
\usepackage{url}
\usepackage{verbatim}
\usepackage{graphicx}
\usepackage{cite}
\usepackage{anyfontsize}

\usepackage{multirow}
\usepackage{booktabs}
\usepackage[ruled,vlined,linesnumbered]{algorithm2e}
\SetKwInput{KwInput}{Input}                
\SetKwInput{KwOutput}{Output}              
\SetKwComment{Comment}{$\triangleright$\ }{}
\usepackage{xcolor, pifont}

\begin{document}

\title{Wireless Traffic Prediction with Large Language Model}

\author{Chuanting~Zhang,~\IEEEmembership{Senior Member,~IEEE,}
        Haixia Zhang,~\IEEEmembership{Senior Member,~IEEE,}
        Jingping~Qiao,~\IEEEmembership{Member,~IEEE,}
        Zongzhang Li,
        and~Mohamed-Slim Alouini,~\IEEEmembership{Fellow,~IEEE}%
\thanks{This work was supported in part by NSFC under No. 62401338, in part by the Joint Funds of the NSFC under Grant No. U22A2003, in part by the Shandong Province Excellent Youth Science Fund Project (Overseas) under Grant No. 2024HWYQ-028, and by The Fundamental Research Funds of Shandong University.}
\thanks{
Chuanting~Zhang and Haixia~Zhang are with the Institute of Intelligent Communication Technologies, and Shandong Key Laboratory of Intelligent Communication and Sensing-Computing Integration, Shandong University, Jinan 250100, China (E-mail: chuanting.zhang@sdu.edu.cn, haixia.zhang@sdu.edu.cn).}
\thanks{
Jingping Qiao is with the School of Information Science and Engineering, Shandong Normal University, Jinan, 250358, China (E-mail: jingpingqiao@sdnu.edu.cn).
}
\thanks{
Zongzhang Li is with China Mobile Communications Group Shandong Co., Ltd, Jinan, 250001, China (E-mail: lizongzhang@sd.chinamobile.com).
}
\thanks{Mohamed-Slim Alouini is with Computer, Electrical and Mathematical Science and Engineering Division, King Abdullah University of Science and Technology (KAUST), Thuwal 23955-6900, Saudi Arabia (E-mail: slim.alouini@kaust.edu.sa).}
}


\maketitle

\begin{abstract}
The growing demand for intelligent, adaptive resource management in next-generation wireless networks has underscored the importance of accurate and scalable wireless traffic prediction. While recent advancements in deep learning and foundation models such as large language models (LLMs) have demonstrated promising forecasting capabilities, they largely overlook the spatial dependencies inherent in city-scale traffic dynamics. In this paper, we propose TIDES (\underline{T}raffic \underline{I}ntelligence with \underline{D}eepSeek-\underline{E}nhanced \underline{S}patial-temporal prediction), a novel LLM-based framework that captures spatial-temporal correlations for urban wireless traffic prediction. TIDES first identifies heterogeneous traffic patterns across regions through a clustering mechanism and trains personalized models for each region to balance generalization and specialization. To bridge the domain gap between numerical traffic data and language-based models, we introduce a prompt engineering scheme that embeds statistical traffic features as structured inputs. Furthermore, we design a DeepSeek module that enables spatial alignment via cross-domain attention, allowing the LLM to leverage information from spatially related regions. By fine-tuning only lightweight components while freezing core LLM layers, TIDES achieves efficient adaptation to domain-specific patterns without incurring excessive training overhead. Extensive experiments on real-world cellular traffic datasets demonstrate that TIDES significantly outperforms state-of-the-art baselines in both prediction accuracy and robustness. Our results indicate that integrating spatial awareness into LLM-based predictors is the key to unlocking scalable and intelligent network management in future 6G systems.
\end{abstract}


\section{Introduction}\label{sec:introduction}

Wireless networks face a data explosion. Between 2014 and 2019, global mobile traffic surged tenfold, and by 2023, nearly 30 billion devices joined the network. This growth, driven by data-intensive applications and emerging industries, such as the Internet of Things, augmented reality, extended reality, and connected vehicles, places significant strain on existing network capacities \cite{wang2023comct,Naboulsi2016comst,zhang2021wcl}.

Traditional networks, designed for static conditions, struggle with today's highly dynamic and fluctuating traffic, resulting in poor resource utilization and degraded user experience. With 5G nearing its limits, intelligent AI-based control methods are urgently needed \cite{lei2025jsac,lei2025network}. Indeed, the next generation, 6G, is expected to integrate AI deeply into network management, paving the way for networks that autonomously optimize themselves \cite{jiang2025comprehensive,cui_overview_2025,dong2025commag,chen2024mwc,li2017mwc}. In this context, large language models (LLMs) \cite{radford2018improving} are anticipated to play a pivotal role in advancing 6G communications by introducing semantic understanding, reasoning, and knowledge generalization capabilities. By effectively interpreting multimodal network data and inferring user intents, LLMs can enable intent-aware network control, predictive resource management, and cross-domain knowledge transfer across heterogeneous environments.

At the heart of this revolution lies wireless traffic prediction \cite{wang2019tmc,zhang2018stdensenet,xu2019jsac,xiao2025cellular,SHANG2024112333,zhang2025jsactides}, i.e., the ability to foresee traffic changes, enabling networks to act proactively. Accurate predictions allow mobile network operators to dynamically adjust resources, balance loads, and reduce energy consumption, directly enhancing efficiency and user experience \cite{zhang2021fedda,zhang2024tccn,gao2024icct}.

To solve wireless traffic prediction problems\cite{sun2025wasamgpt,zhang2025cit,gao2025math,gao2024icic,shen2021}, researchers initially modeled wireless traffic using simple statistical methods like Autoregressive Integrated Moving Average (ARIMA) and its variants \cite{Shu2005}. These models were adequate for stable conditions but failed to capture today's complex, nonlinear traffic patterns \cite{de200625,zhang2021complexity}. Machine learning improved predictions, yet required extensive manual feature selection \cite{Sapankevych2009cim}.

The breakthrough came with deep learning \cite{lecun2015deep,zhang2016smartdata,zhang2015cbdcom}, or deep neural networks, which learn complex spatiotemporal patterns directly from raw data, significantly surpassing previous methods. Techniques like convolutional neural networks (CNN) \cite{he2016deep} for spatial relationships, long-short term memory (LSTM) networks for temporal patterns, and graph neural networks \cite{kipf2016semi} demonstrated impressive accuracy. Federated and transfer learning approaches further enabled data sharing across regions while preserving privacy \cite{chenxi2021globecom,zhang2022fedmeta}.

\begin{figure*}[!ht]
\centering
\includegraphics[width=0.9\textwidth]{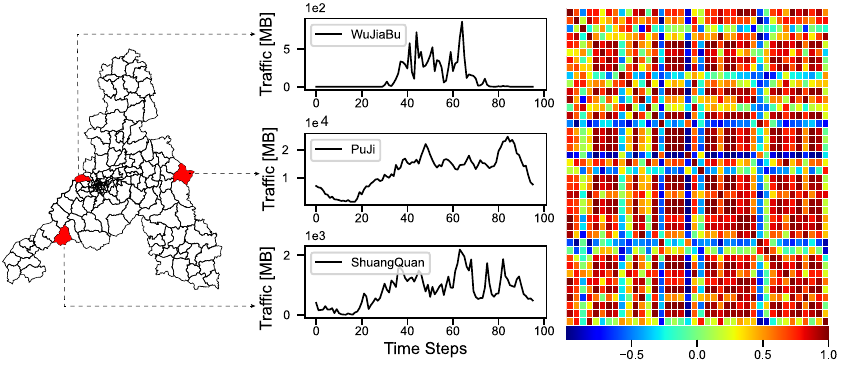}
\caption{Traffic patterns vary greatly among different places. Left: City boundary and the three selected areas, i.e., WuJiaBu, PuJi, and ShuangQuan. Middle: The temporal traffic dynamics of WuJiaBu, PuJi, and ShuangQuan. Right: Pearson correlation coefficient of forty randomly selected areas from the city.}
\label{motivated_example}
\end{figure*}

Recently, TrafficLLM \cite{hu2024self} and Time-LLM \cite{jin2023time} emerged as promising frameworks, leveraging their powerful sequence-learning abilities through prompt-based predictions. These models learn from minimal data, making them highly adaptive and efficient. However, existing approaches primarily focus on temporal predictions for isolated cells using a single trained LLM model, neglecting critical spatial interactions between neighboring cells.
In reality, wireless traffic shows not only temporal patterns, but also inherently spatial patterns. Fig. \ref{motivated_example} serves as a motivating example illustrating why a single, universal prediction model fails to capture diverse traffic patterns across a city. The left highlights three distinct urban areas, i.e., WuJiaBu, PuJi, and ShuangQuan, marked in red on the city map. The middle depicts their dramatically different temporal traffic behaviors, emphasizing the uniqueness of each region's traffic dynamics. The right further demonstrates this variability through a measure of pattern similarity, clearly revealing heterogeneous spatial correlations. This figure indicates that neighboring cells influence each other's traffic due to user mobility and overlapping coverage. Ignoring these spatial dependencies limits predictive accuracy and scalability. Current LLM-based methods fail to effectively integrate spatial awareness, creating an urgent need for solutions that capture both temporal dynamics and spatial context simultaneously.

To fill this gap, we propose TIDES, which is Traffic Intelligence with DeepSeek-enhanced Spatial-temporal prediction —a novel prediction framework that uniquely combines the strengths of large AI models with region-specific spatial intelligence. Specifically, our contributions can be summarized as:
\begin{itemize}
    \item Region-aware modeling. TIDES groups different regions by traffic similarity, then fine-tunes a common foundation model for each group. This approach balances generalization across a network and specialization for individual regions, enhancing prediction accuracy without excessive complexity.
    \item Prompt-based traffic representation. We convert traffic data into structured natural-language prompts that LLMs easily understand, leveraging their innate capacity to detect patterns and generalize predictions with minimal training.
    \item Spatial–temporal alignment with DeepSeek. TIDES incorporates a special attention mechanism that enables LLMs to consider traffic data from neighboring cells, effectively aligning spatial context within predictions. More importantly, this alignment is efficient, as it avoids retraining large models entirely by fine-tuning just a few parameters.
\end{itemize}

This innovative design allows TIDES to capture citywide spatial dynamics effectively, significantly improving prediction accuracy. Our results validate TIDES as a scalable, spatially intelligent predictor capable of generalizing across diverse urban environments. Ultimately, TIDES advances network intelligence toward fully autonomous, proactive management, which is essential for next-generation 6G networks.

\section{Related Works}
Wireless traffic prediction has been approached through a progression of methodologies, from classical time-series analysis \cite{campos2023lightts} to modern deep learning and, most recently, large language models. This section reviews key representative works in wireless traffic prediction, organized by methodology: traditional statistical models, shallow machine learning, deep learning approaches, and emerging LLM techniques. 

\subsection{Traditional Statistical Methods}
Early research on forecasting cellular traffic mainly used statistical models like ARIMA to identify patterns over time \cite{Shu2005}. While ARIMA can track daily or weekly trends, it often fails to handle wireless traffic's unpredictable ups and downs. To tackle these issues, researchers have looked at combining ARIMA with GARCH (Generalized Autoregressive Conditional Heteroskedasticity) \cite{zhou2006traffic} and other methods like $\alpha$-stable distributions \cite{li2019tcomm} and entropy measures \cite{li2014mcom} to better capture variability. Although these techniques can provide clear insights in stable conditions, their accuracy decreases with complex trends or sudden changes. This has led many to look for data-driven learning methods to overcome the limits of traditional statistical techniques.
\subsection{Shallow Machine Learning Algorithms}
With the rise of data mining techniques, machine learning (ML) models were applied to wireless traffic data as a next step beyond basic time-series methods. Shallow ML models such as support vector regression \cite{chen2022traffic}, linear regression, and decision trees have shown competitive accuracy on certain wireless traffic prediction tasks. These models can capture some nonlinearity in the data and are relatively fast to train, though they typically require careful feature engineering, e.g., using lagged traffic values and time-of-day indicators, to handle temporal patterns. Ensemble approaches, such as random forests and gradient boosting machines \cite{stepanov2020applying}, have also been used to enhance prediction robustness. For example, a framework that first performs data reduction via clustering, then selects an optimal learning algorithm for each cluster’s traffic pattern was proposed in \cite{Nashaat2024ACCESS}. 
This adaptive ensemble approach achieved high accuracy while significantly cutting computational cost compared to baseline methods. 
These results suggest that, with appropriate preprocessing, non-deep learning models can still provide effective short-term forecasts in wireless networks. 
Nonetheless, as wireless traffic patterns grew more complex, the trend in recent years has shifted toward deep learning, which can automatically capture intricate spatiotemporal dependencies. 
\subsection{Deep Learning Models}
In recent years, deep learning has established itself as the leading approach for wireless traffic prediction \cite{zeng2023transformers}. Unlike shallow models, deep neural networks effectively learn hierarchical features from raw data, making them adept at modeling the complex temporal dynamics and spatial correlations in mobile networks. Early implementations included LSTM networks, which demonstrated greater accuracy than ARIMA models in base station traffic prediction \cite{Wang2017,Qiu2018}. While recurrent models excel at capturing temporal sequences, spatial dependencies are equally crucial in urban traffic contexts, prompting researchers to combine temporal and spatial modeling techniques. Notably, some studies have integrated CNNs with LSTMs; for example, convolutional LSTM networks with attention mechanisms have been employed to learn joint spatial-temporal features from multiple cell sites \cite{zhang2018stdensenet}.

Another research trend involves incorporating external factors and auxiliary data into deep learning frameworks. The STCNet model \cite{zhang2019stcnet} effectively utilizes diverse external features, including demographic data and weather conditions, to enhance prediction accuracy. By leveraging knowledge from related domains, this model improves the anticipation of traffic surges driven by external events, employing deep transfer learning techniques. 
Graph neural networks (GNNs) have gained traction for modeling cellular networks, treating each cell tower as a node within a graph. For instance, an adaptive graph convolutional recurrent network (AGCRN) effectively learns both traffic patterns and the relationships among cell sites \cite{bai2020adaptive}. Such models provide insights into how traffic fluctuations at one base station can affect another.

Beyond traditional supervised learning, researchers are exploring meta-learning to enhance model generalization. Bayesian meta-learning has been applied to rapidly adapt prediction models to new areas with limited data \cite{wang2024tmc,li2023twc}, addressing distribution shifts over time. While much deep learning research has focused on short-term predictions, some recent work targets long-term predictions. The GCformer model \cite{zhao2023cikm}, which combines graph neural networks with Transformer-based sequence modeling \cite{nie2022time,liu2023itransformer,wu2021autoformer}, aims to predict multi-step 5G base station traffic over longer horizons, a crucial task for capacity planning. Despite the propensity for prediction errors to increase with longer forecast horizons, short-term predictions remain the primary focus for real-time network optimization, highlighting deep learning’s transformative role in wireless traffic prediction.

\subsection{LLMs Models and Frameworks}
Recent advancements in wireless traffic prediction involve using LLMs and foundational models. Researchers are reimagining traffic prediction as a language modeling task, converting numerical time-series data into textual prompts for LLMs. \textit{Shokouhi} and \textit{Wong} pioneered this by encoding historical traffic levels and data from similar-pattern cells into natural language sentences \cite{shokouhi2024globecom}. They fine-tuned LLMs like BART and BigBird for forecasting, resulting in a prediction model that outperformed a graph neural network model.
Additionally, researchers are generating synthetic traffic data using LLMs. \textit{Dual et. al} proposed an approach to produce realistic traffic patterns tailored to specific parameters \cite{duan2024globecom}. Synthetic data can supplement real datasets, addressing issues of scarcity and privacy. Initial results suggest that LLMs can generate traffic sequences that maintain the statistical properties of actual network loads.

Another innovative method, TrafficLLM \cite{hu2024self}, uses a self-refining generative approach. Here, the pre-trained model makes predictions based on a few examples, generates feedback on any inaccuracies, and refines its predictions without gradient-based training. This method improved prediction accuracy greatly on real 5G datasets without task-specific tuning, making it appealing for deployment.

In summary, LLMs are proving to be effective tools for wireless traffic prediction, capable of integrating diverse contextual information through fine-tuning or strategic prompting.

\begin{figure*}[!ht]
\centering
\includegraphics[width=1.0\textwidth]{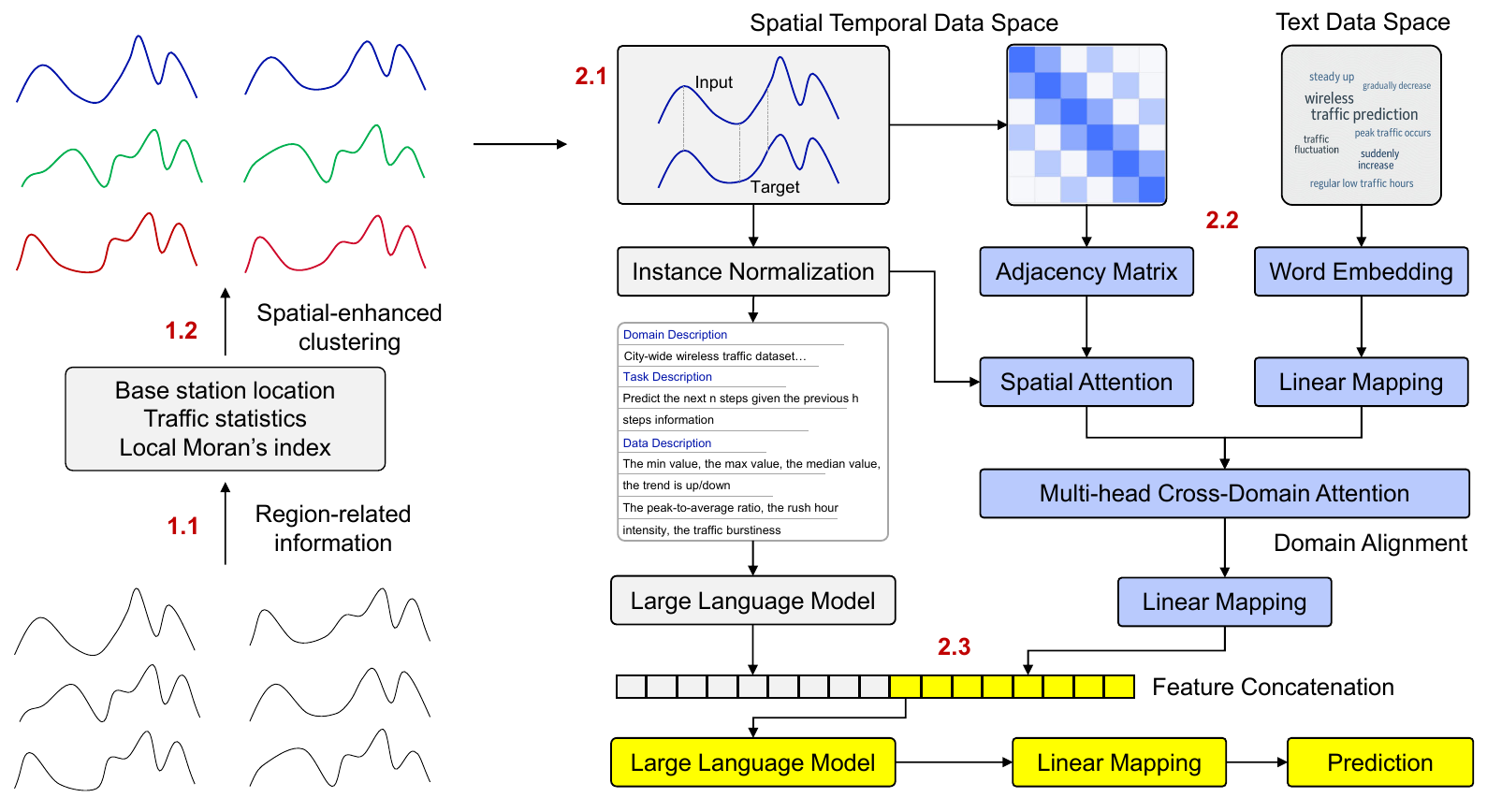}
\caption{Wireless traffic prediction with a two-phase process. The first phase (left) involves a spatial-aware clustering process, and the second phase (right) utilizes the TIDES framework, which performs LLM-driven spatial-temporal learning.}
\label{sys_arch}
\end{figure*}

\section{Problem Formulation and Preliminaries}
We first formalize our wireless traffic prediction problem and then introduce some preliminaries on LLM.
\subsection{Problem Formulation}
Let us denote the wireless traffic dataset as a spatial-temporal matrix composed of traffic records across $R$ spatial regions (e.g., city zones or base station clusters). For each region $i \in \{1, 2, \ldots, R\}$, we define a time-series traffic signal $\mathbf{x}_i = \{x_i^1, x_i^2, \ldots, x_i^T\}$, where $x_i^t \in \mathbb{R}$ represents the downlink traffic volume observed at time step $t$, and $T$ is the total number of time steps in the dataset.

Given a historical window size $H$, and a prediction horizon $P$, the objective is to learn a mapping function $\mathcal{F}$ such that:
\begin{align}
\begin{split}
    \{\mathbf{\hat{y}}_1,\mathbf{\hat{y}}_2,\cdots,\mathbf{\hat{y}}_R\} &= \mathcal{F}(\mathbf{X}^{t-H+1:t}, \mathcal{G}; \mathbf{W}) \\
    &= \{\mathbf{x}_1^{t+1:t+P}, \mathbf{x}_2^{t+1:t+P},\cdots, \mathbf{x}_R^{t+1:t+P}\}
\end{split}
\end{align}
where $\hat{\mathbf{y}}_i$ denotes the predicted traffic values for region $i$ over the next $P$ time steps, based on the past $H$ observations $\mathbf{X}^{t-H+1:t}$. $\mathbf{W}$ represents the weights of the mapping function $\mathcal{F}$.
Additionally, $\mathcal{G} = (\mathcal{V}, \mathcal{E})$ is a spatial graph, in which each node $v_i \in \mathcal{V}$ corresponds to region $i$, and edges $(v_i, v_{j}) \in \mathcal{E}$ reflect spatial proximity or statistical similarity between regions $i$ and $j$. The goal is to learn a spatial-temporal prediction model that jointly leverages $\{ \mathbf{x}_i\}_{i=1}^R$ and the graph structure $\mathcal{G}$ to enhance forecasting accuracy.

\subsection{Preliminaries on Large Language Models}
LLMs, such as GPT and DeepSeek, are trained to model the distribution over sequences by learning conditional probabilities of the form:
\begin{equation}
P(x_1,x_2,\cdots,x_n)=\prod_{t=1}^nP(x_t|x_1,x_2,\cdots,x_{t-1})
\end{equation}
where $x_t$ denotes the $t$-th token in an input sequence. These models are typically built upon transformer decoders that rely on self-attention mechanisms to capture long-range dependencies.
\subsubsection{GPT}
GPT is an autoregressive transformer model that generates tokens one by one, predicting each token conditioned on its left-side context. Each decoder block in GPT consists of the following sub-layers: multi-head self-attention (MHA), add \& layer normalization, position-wise feed-forward network (FFN), and add \& layer normalization. The above structure is repeated $L$ times in a stack, forming the deep architecture of GPT.

MHA allows the model to attend to all previous tokens but not future ones, preserving causality. Each attention head computes
\begin{equation}
    \text{Attention}_i^\text{MHA} = \text{softmax}(\frac{\mathbf{Q}_i\mathbf{K}_i^{\top}}{\sqrt{d_K}})\mathbf{V}_i, \label{att_mha}
\end{equation}
where $\mathbf{Q}_i, \mathbf{K}_i, \mathbf{V}_i$ are projections of the input for head $i$, and $d_K$ is the dimensionality of the key vectors. Multiple heads are computed in parallel, that is, 
\begin{align}
    \mathbf{Q}_i&=\mathbf{H}^{(l-1)}\mathbf{W}_{Q,i}, \label{att_q} \\ \mathbf{K}_i&=\mathbf{H}^{(l-1)}\mathbf{W}_{K,i}, \label{att_k} \\ \mathbf{V}_i&=\mathbf{H}^{(l-1)}\mathbf{W}_{V,i} \label{att_v}.
\end{align}
Note that $\mathbf{H}^{(0)}=\mathbf{E}_{token}+\mathbf{P}$ denotes the input including token embedding from input $\mathbf{x}$ and the corresponding positional embedding. Then the outputs from all $h$ heads are concatenated and projected, which can be described as
\begin{equation}
    \text{MHA}(\mathbf{H})=(\Vert_{i=1}^h \text{Attention}_i) \mathbf{W}_O
\end{equation}
This attention mechanism is crucial for contextual representation learning, allowing each token to gather information from relevant past tokens.

After attention, the representation is passed through a layer normalization yet with a residual connection of $\mathbf{H}^{(l-1)}$,
\begin{equation}
    \mathbf{Z}=\text{LayerNorm}(\mathbf{H}^{(l-1)}+\text{MHA}(\mathbf{H}^{(l-1)})).
\end{equation}
Then, $\mathbf{Z}$ is mapped with a position-wise feed-forward network, shared across all positions:
\begin{equation}
\text{FFN}(\mathbf{Z})=\text{GELU}(\mathbf{Z}\mathbf{W}_1)\mathbf{W}_2,
\end{equation}
where $\mathbf{W}_1, \mathbf{W}_2$ are learnable matrices with typically 4 times expansion in hidden size and GELU is the activation function.
After that, the representation is again passed through a layer normalization by the following operation:
\begin{align}
    \mathbf{H}^{(l)}&=\text{LayerNorm}(\mathbf{Z}+\text{FFN}(\mathbf{Z})).
\end{align}
GPT uses a pre-layer normalization operation, that is, it places the normalization before attention and FFN in practice.

After the final decoder block $L$, the model computes token logits using a tied projection matrix ($\mathbf{E}$) by
\begin{equation}
    \mathbf{Y}=\mathbf{H}^{(L)} \mathbf{E}^{\top}.
\end{equation}

\subsubsection{DeepSeek Foundation Model}
DeepSeek is an open-source foundation model designed for general-purpose tasks, including reasoning and cross-modal learning. It extends the transformer architecture with a significantly larger model size and specialized training corpus, including domain-specific knowledge such as structured data and tabular formats. One major difference between DeepSeek and GPT is that the former adopts multi-query attention (MQA) instead of multi-head attention.

In MQA, queries ($\mathbf{Q}$) remain head-specific, but keys ($\mathbf{K}$) and values ($\mathbf{V}$) are shared across all heads. This reduces memory footprint and improves inference speed, especially beneficial in deployment scenarios or large-scale autoregressive decoding. The attention score calculation in MQA can be expressed as
\begin{equation}\label{att_mqa}
    \text{Attention}_i^\text{MQA} = \text{softmax}(\frac{\mathbf{Q}_i\mathbf{K}^{\top}}{\sqrt{d_K}})\mathbf{V}.
\end{equation}

In the context of traffic prediction, the traffic sequence can be serialized as a pseudo-language sequence, where traffic values, region identifiers, and timestamps are embedded into tokens. The model is then fine-tuned or prompted to forecast future traffic values based on historical traffic token streams.

\section{Our Proposed Method}
We propose a novel method for wireless traffic prediction that uniquely integrates spatial-temporal analysis, statistical traffic characterization, and LLMs in a cohesive architecture. Our approach addresses the fundamental limitations of traditional forecasting methods by incorporating domain knowledge from both spatial and linguistic perspectives.
The workflow of our method is displayed in Fig. \ref{sys_arch}, which is actually a two-phase process with the first phase being a spatial-aware region clustering and the second phase being an LLM-based prediction model. Particularly, we adopt DeepSeek as our base LLM model.

In the first step of phase 1, region-related information is extracted, including geographic location, traffic statistics, and spatial autocorrelation metrics. Then, in the second step of phase 1, clustering is applied using these statistics, grouping similar regions together. 
The second phase operates on each cluster generated by the first phase and performs modeling training. The training process consists mainly of three components: prompt engineering, spatial-attention enhanced domain alignment, and LLM-based feature representation learning. These four components are marked as 2.1, 2.2, and 2.3 in Fig. \ref{sys_arch}. We detailed each step in the following sections.

\subsection{Spatial-Aware Region Clustering}
Our framework incorporates a novel clustering methodology that effectively leverages both traffic patterns and spatial autocorrelation to group base stations with similar characteristics. This approach enables more accurate traffic prediction by capturing the inherent spatial dependencies in wireless networks.

\subsubsection{Region-related information extraction}
For each region $i$, we construct a multidimensional feature vector $\mathbf{e}_i$ that encapsulates both geographic and traffic characteristics:

\begin{equation}\label{region_feature}
\mathbf{e}_i = [\phi_i, \lambda_i, x_i^\text{MEAN}, x_i^\text{AM}, x_i^\text{PM}, x_i^\text{NIGHT}, I_i],
\end{equation}
where:
\begin{itemize}
    \item $\phi_i, \lambda_i$ represent the latitude and longitude coordinates.
    \item $x_i^\text{MEAN}$ is a normalized transformation of mean traffic volume.
    \item $x_i^\text{AM}, x_i^\text{PM}, x_i^\text{NIGHT}$ represent traffic during morning peak, evening peak, and night periods, respectively.
    \item $I_i$ is the local Moran's I statistic.
\end{itemize}

This multidimensional representation captures both the geographic location and the temporal traffic patterns of each base station, providing a rich foundation for clustering.

To obtain $I_i$, we first measure the physical distance for any given region pair $i$ and $j$ using the Haversine formula, accounting for Earth's curvature:
\begin{equation}
d_{ij} = 2r \cdot \arcsin\left(\sqrt{\sin^2\left(\frac{\Delta\phi}{2}\right) + \cos\phi_i\cos\phi_j\sin^2\left(\frac{\Delta\lambda}{2}\right)}\right),
\end{equation}
where $r$ is Earth's radius, $\phi_i$ and $\phi_j$ are the latitudes of regions $i$ and $j$ in radians. $\Delta\phi$ is the difference in latitude, $\Delta\lambda$ is the difference in longitude. Be noted that traffic similarity can be an alternative for measuring distance when latitude and longitude coordinates are unavailable.

We construct a spatial weight matrix $\mathbf{A}$ based on $k$-nearest neighbors, where the weight $a_{ij}$ is nonzero if region $j$ is among the $k$ nearest neighbors of region $i$:

\begin{equation}
a_{ij} = \begin{cases}
1/d_{ij} & \text{if}\ j \in \mathcal{N}_k(i) \\
0 & \text{otherwise}
\end{cases},
\end{equation}
where $d_{ij}$ is the Haversine distance between regions $i$ and $j$, and $\mathcal{N}_k(i)$ is the set of $k$ nearest neighbors to region $i$.

For each region, the local Moran's I statistic is calculated by:

\begin{equation}
I_i = z_i \sum_{j} a_{ij} z_j,
\end{equation}
where $z_i = (x_i^t - x_i^\text{MEAN})/x_i^\text{STD}$ is the standardized traffic value at region $i$, with $x_i^\text{STD}$ being the standard deviation of traffic values of that region. We use Moran's I to guide clustering so that regions grouped together share not only similar traffic distributions but also significant spatial interactions.

\subsubsection{Enhanced K-means clustering}

We implement an enhanced K-means clustering algorithm that incorporates the spatial autocorrelation measures. Given a set of regions with feature vectors $\{\mathbf{e}_1, \mathbf{e}_2, \ldots, \mathbf{e}_R\}$, we aim to partition them into $K$ clusters $\mathcal{C} = \{C_1, C_2, \ldots, C_K\}$. To stabilize the clustering, these feature vectors are normalized to have zero means and unit variance.

The objective function is defined as:

\begin{equation}
\min_{\mathcal{C}} \sum_{k=1}^{K} \sum_{\mathbf{e}_i \in C_k} \|\mathbf{e}_i - \boldsymbol{\mu}_k\|^2,
\end{equation}
where $\boldsymbol{\mu}_k$ is the centroid of cluster $C_k$.

\subsection{TIDES Framework}
Our proposed TIDES framework mainly consists of three components, i.e., prompt engineering, spatial attention-enhanced domain alignment, and LLM-based feature representation and output projection.
\subsubsection{Prompt engineering}
Before we proceed to the next step, we first normalize each input traffic data $\mathbf{x}_i$ to have zero mean and unit standard deviation by using reversible instance normalization (RevIN). We do this because RevIN can not only stabilize training by ensuring inputs fall within a consistent range, preventing gradient explosion or vanishing, but also enables the model to focus on relative patterns rather than absolute magnitudes, which vary significantly across different regions and time periods. We still use $\mathbf{x}_i$ to denote the normalized version of it to avoid notation redundancy.

For the input wireless traffic data $\mathbf{x}_i$, we extract the following features: fundamental statistics, trend, time of day patterns, and domain-specific metrics.

\textbf{Fundamental Statistical Measures}: The minimum, maximum, median, mean, and standard deviation values of $\mathbf{x}$, which are described as
\begin{equation}
    \mathcal{P}_\text{STAT}=\{x_i^\text{MIN}, x_i^\text{MAX}, x_i^\text{MED}, x_i^\text{MEAN}, x_i^\text{STD}\}.
\end{equation}
Basic statistics provide fundamental distribution characteristics of the time series data. These metrics help the LLM understand the scale, central tendency, and dispersion of traffic values, establishing baseline expectations for forecasting. They are essential for capturing the static properties of the traffic pattern and help the model calibrate its predictions within reasonable bounds.

\textbf{Trend Analysis}: The overall trend is calculated as the sum of first-order differences, 
\begin{equation}
\mathcal{P}_\text{TND}=\{x_i^\text{TND}\}.
\end{equation}
The trend direction $x_i^{\text{TND}}=\sum_t(\mathbf{x}^{t+1}-\mathbf{x}^t)$ is marked as upward if $x_i^\text{TND} \geq 0$ else as downward. Trend information is crucial for capturing the directional momentum of traffic patterns. Wireless traffic often exhibits persistent trends due to evolving user behaviors, population movements, or network capacity changes. By explicitly encoding the trend direction, we help the LLM distinguish between growth and decline phases, which is particularly important for forecasting during transition periods or when predicting longer horizons where trend continuation or reversal significantly impacts accuracy.

\textbf{Time of Day Patterns}: The wireless traffic of different regions exhibits strong daily patterns, including morning peaks, evening peaks, and night peaks. We calculate these time-of-day patterns
\begin{equation}
    \mathcal{P}_\text{TOD}=\{ x_i^\text{AM}, x_i^\text{PM}, x_i^\text{NIGHT}, x_i^\text{NonRush}\},
\end{equation}
where $x_i^\text{NonRush}$ means the traffic volume at non-rush hours.
Time-of-day patterns directly capture human activity cycles that drive wireless network usage. By explicitly computing statistics for morning commutes, evening returns, and overnight periods, we encode domain knowledge about human mobility and communication patterns. These features are particularly important because they represent predictable, recurrent behaviors that strongly influence traffic volumes at base stations across urban environments.

\textbf{Domain Specific Metrics}: We then extract wireless-specific statistical features including peak-to-average ratio $x_i^\text{PAR}$, rush-hour intensity $x_i^\text{RHI}$, morning-evening ratio $x_i^\text{MER}$, traffic burstiness $x_i^\text{BURST}$, and volatility change $x_i^\text{VC}$. These features can be described as

\begin{equation}
    \mathcal{P}_\text{DSM}=\begin{cases}
        x_i^\text{PAR}&=x_i^\text{MAX}/({x_i^\text{MAX} + \epsilon}), \\
        x_i^\text{RHI}&=(x_i^\text{AM}+x_i^\text{PM})/(2\cdot{x_i^\text{NonRush} + \epsilon}), \\
        x_i^\text{MER}&=x_i^\text{AM}/(x_i^\text{PM}+\epsilon), \\
        x_i^\text{BURST}&=x_i^\text{STD}/(x_i^\text{MEAN}+\epsilon),\\
        x_i^\text{VC}&=x_{i,\text{Second}}^\text{STD} / (x_{i,\text{First}}^\text{STD}+\epsilon)
    \end{cases}
\end{equation}
where $x_{i,\text{First}}^\text{STD}$ and $x_{i,\text{Second}}^\text{STD}$ denotes the standard deviation of the first half and the second half of $\mathbf{x}_i$, respectively. $\epsilon$ is a small value that prevents the occurrence of zero division.

The feature $x_i^\text{PAR}$ captures traffic spikes relative to the average load, highlighting burstiness that impacts capacity planning. $x_i^\text{RHI}$ quantifies the difference between peak and off-peak traffic volumes, reflecting location-specific usage patterns. The morning-evening ratio characterizes asymmetry between morning and evening traffic, aiding in area-specific forecasting. Burstiness measures overall traffic irregularity, distinguishing stable from sporadic usage, while volatility change captures shifts in variability, useful for identifying emerging patterns or instabilities. Collectively, these features allow models to adapt to diverse traffic characteristics, enhancing forecasting accuracy under dynamic conditions.

To fully utilize these descriptors, we construct a natural language prompt, denoted as $s_\mathcal{P}$, by concatenating all components as shown in Fig. \ref{sys_arch}. This multi-feature prompt transforms raw traffic data into a structured, interpretable format, enabling LLMs to better leverage their reasoning and pattern recognition capabilities.
The representation after feeding the prompt into LLM can be written as 
\begin{equation}
\mathbf{H}^\text{Prompt} = \text{LLM}(s_\mathcal{P}).
\end{equation}

\subsubsection{Domain Alignment}

Our model incorporates a spatial attention (SA) mechanism that explicitly captures the topological structure of communication networks, extending beyond conventional temporal-focused designs. For each cluster, we construct an adjacency matrix based on regional proximity, avoiding the direct use of raw distance information. Instead, TIDES applies a graph shift operation to the adjacency matrix to ensure numerical stability, as the raw form may induce gradient explosion or vanishing due to widely spread eigenvalues. Additionally, the Laplacian-based graph shift operation emphasizes local structural patterns, enabling the model to identify regional traffic anomalies and localized dynamics more effectively.
The symmetric normalized Laplacian transformation can be described as
\begin{equation}
\mathbf{L}_{sym} = \mathbf{I} - \mathbf{D}^{-1/2}\mathbf{A}\mathbf{D}^{-1/2}
\end{equation}
where $\mathbf{D}$ is the degree matrix with diagonal elements $D_{ii} = \sum_j A_{ij}$, and $\mathbf{I}$ is the identity matrix.

After obtaining $\mathbf{L}_{sym}$, we proceed to model spatial correlations using a multi-head attention mechanism. For the $i$-th head, we first map the input $\mathbf{X}$, the matrix version of $\mathbf{x}$, into query, key, and value representations. Specifically, the transformation can be written as
\begin{align}
\mathbf{Q}_i^\text{SA} = \mathbf{X}\mathbf{W}^\text{SA}_{Q,i},
\mathbf{K}_i^\text{SA} = \mathbf{X}\mathbf{W}^\text{SA}_{K,i},
\mathbf{V}_i^\text{SA} = \mathbf{X}\mathbf{W}^\text{SA}_{V, i},
\end{align}
where $\mathbf{W}^\text{SA}_{Q,i}, \mathbf{W}^\text{SA}_{K,i}, \mathbf{W}^\text{SA}_{V, i}$ are learnable projection parameters.

The outputs of multiple heads are then concatenated through $\mathbf{Q}^\text{SA} = ||_{i=1}^h \mathbf{Q}_i^\text{SA}$, $\mathbf{K}^\text{SA} = ||_{i=1}^h \mathbf{K}_i^\text{SA}$, and $\mathbf{V}^\text{SA} = ||_{i=1}^h \mathbf{V}_i^\text{SA}$, respectively.
This multi-head attention offers several benefits. It not only allows the model to attend to information from different representation subspaces simultaneously, but also enables each attention head to focus on different types of spatial relationships between regions. For instance, some heads might capture geographic proximity effects, others might detect functional relationships (e.g., stations serving similar types of venues like shopping centers or transport hubs), and additional heads could specialize in temporal synchronization patterns.

Next, we compute the attention scores and incorporate spatial constraints as follows:
\begin{equation}
\mathbf{H}^\text{SA} = \text{softmax}\{\frac{\mathbf{Q}^\text{SA}(\mathbf{K}^\text{SA})^\top}{\sqrt{d_K}} + \mathbf{M}\}\mathbf{V}^\text{SA},
\end{equation}
where $\mathbf{M}=(1-\mathbf{L}_{sym})\cdot (-1000)$ serves as a masking operation on the adjacency matrix. The motivations for this operation are twofold: 1) In reality, not all base stations directly influence each other. Signal interference and handovers usually occur only between neighboring stations; 2) By masking non-connected pairs, we ensure that the attention mechanism only considers realistic relationships based on the actual network topology.

Each trained large language model contains an internal vector representation for each word. It is based on this vector representation that the model is able to perform computations on text. Assume that the vocabulary size is $vocab$, and each word is represented by a vector of dimension $d$. Then, the word embeddings can be represented as $\mathbf{E}\in \mathbb{R}^{vocab\times d}$.
For GPT, each word is represented as a 768-dimensional vector. For the DeepSeek model we use, each word is represented by a 4068-dimensional vector.
To reduce computational complexity, we first apply a linear layer to project $\mathbf{E}$ into a lower-dimensional space $\mathbf{E}^\text{Low}$.
Next, we perform domain alignment between the spatiotemporal wireless traffic features $\mathbf{H}^\text{SA}$ and the textual features $\mathbf{E}^\text{Low}$. Specifically, we map $\mathbf{H}^\text{SA}$ into a query matrix, we map $\mathbf{E}^\text{Low}$ into a key matrix and a value matrix, respectively. Then, we perform cross-modal alignment using a multi-head attention mechanism, which enables the model to align feature spaces across regions by learning how information from one domain can improve predictions in another. The multi-head structure further allows the model to capture multiple types of cross-domain relationships simultaneously. The final aligned feature representation can be expressed as
\begin{equation}
\mathbf{H}^\text{DA} = \text{softmax}\{\frac{\mathbf{H}^\text{SA}\mathbf{W}^\text{DA}_Q(\mathbf{E}^\text{Low}\mathbf{W}^\text{DA}_K)^\top}{\sqrt{d_K}}\}(\mathbf{E}^\text{Low}\mathbf{W}^\text{DA}_V).
\end{equation}

\subsection{LLM-based Representation and Output Projection}
After we obtain the prompt representation $\mathbf{H}^\text{Prompt}$ and the domain alignment feature representation $\mathbf{H}^\text{DA}$, we concatenate them and feed them to DeepSeek for spatial-temporal feature representation by
\begin{equation}
\mathbf{H}^\text{Final} =\text{LLM}(\mathbf{H}^\text{Prompt} || \mathbf{H}^\text{DA}).
\end{equation}
Then we perform a final output project using a linear layer and obtain the prediction.
\begin{equation}
   \hat{\mathbf{Y}} = \mathbf{H}^\text{Final}\mathbf{W}^\text{out},
\end{equation}
where $\mathbf{W}^\text{out}$ denotes learnable parameters in the last layer of our TIDES framework.

\section{Experiment Results and Analysis}

In this section, we first briefly introduce the datasets we adopted. Then we give the parameter configurations, evaluation metrics, and baseline methods. Finally, we report the results obtained comprehensively, from city-level overall prediction performance to single region-level prediction performance, including both the spatial and temporal perspectives.

\begin{table*}[ht]
\centering
\caption{Prediction performance comparisons among various algorithms regarding MAE, RMSE, and MAPE across four zones (clusters) of a city. The top performances are highlighted in bold, and the second-best results are indicated in underline.}
\label{tab:my-table}
\renewcommand\arraystretch{1.3}
\resizebox{1.0\textwidth}{!}{%
\begin{tabular}{@{}c|ccc|ccc|ccc|ccc@{}}
\toprule
            & \multicolumn{3}{c|}{Zone A} & \multicolumn{3}{c|}{Zone B} & \multicolumn{3}{c|}{Zone C} & \multicolumn{3}{c}{Zone D} \\ \midrule
Method      & MAE     & RMSE    & MAPE   & MAE     & RMSE    & MAPE   & MAE     & RMSE   & MAPE    & MAE     & RMSE    & MAPE   \\ \midrule
DLinear     & \underline{0.2330}  & \underline{0.3133}  & \underline{2.7904} & 0.4384  & 0.6840  & 2.6294 & \underline{0.2751}  & \underline{0.3804}  & 3.5069 & 0.1492  & 0.1940  & \underline{0.6519} \\
Transformer & 0.3633  & 0.4728  & 5.6137 & 0.5582  & 0.8539  & 3.6093 & 0.3326  & 0.4360  & 3.5011 & 0.2102  & 0.2655  & 1.0232 \\
Autoformer  & 0.5120  & 0.6369  & 4.1527 & 0.6812  & 0.9343  & \underline{2.6243} & 0.6666  & 0.8106  & 2.9052 & 0.5587  & 0.6680  & 1.9279 \\
LightTS     & 0.2638  & 0.3515  & 3.0120 & 0.4586  & 0.7074  & 2.7322 & 0.3273  & 0.4387  & 3.6620 & 0.2212  & 0.2793  & 1.0264 \\
Reformer    & 0.3977  & 0.5076  & 6.3745 & 0.5853  & 0.8845  & 3.4789 & 0.3799  & 0.4941  & 4.1688 & 0.3144  & 0.4101  & 1.7972 \\
TimesNet    & 0.2636  & 0.3508  & 4.1973 & 0.4722  & 0.7203  & 2.6261 & 0.2782  & 0.3860  & \underline{2.8450} & \underline{0.1371}  & \underline{0.1795}  & 0.6848 \\
Time-LLM        & 0.2648  & 0.4350  & 5.2143 & \bf{0.3567}  & \bf{0.5401}  & 2.6623 & 0.3020  & 0.4867  & 2.9110 & 0.1914  & 0.2550  & 1.2075 \\ \midrule
TIDES       & \bf{0.2193}  & \bf{0.2958}  & \bf{2.7481} & \underline{0.4208}  & \underline{0.6459}  & \bf{2.4827} & \bf{0.2657}  & \bf{0.3621}  & \bf{2.8358} & \bf{0.1356}  & \bf{0.1746}  & \bf{0.5150} \\ \bottomrule
\end{tabular}%
}
\end{table*}

\subsection{Datasets and Evaluation Metrics}
We validate the TIDES framework on a real-world dataset. The dataset is collected from a city-level 4G base station network of a telecom operator in China. The original dataset contains downlink traffic records from approximately 19,000 sectors, covering the period from July 28 to August 25, 2024—a total of four weeks. The data is recorded at a 15-minute interval, meaning traffic load is logged every 15 minutes. We aggregate the sector-level traffic into regional-level traffic, on which we conduct model training and prediction. Note that the whole city is clustered into $K=4$ clusters, and we denote them as zones in the following.
The following evaluation metrics are used to assess the prediction performance:
\begin{align}
    \text{MAE} &= \frac{1}{T}\sum_{t}|y_t - \hat{y_t}|,\\
    \text{RMSE} &= \sqrt{\frac{1}{T}\sum_{t}(y_t - \hat{y_t})^2},\\
    \text{MAPE} &= \frac{100}{T}\sum_{t}|\frac{y_t - \hat{y_t}}{y_t}|,
\end{align}
where $y_t$ denotes the ground truth value at time step $t$, $\hat{y_t}$ be the corresponding predicted value, and $T$ the total number of prediction steps.

\subsection{Baseline Methods}
To comprehensively evaluate the performance of our proposed method, we compare it against a diverse set of state-of-the-art time series forecasting models. These baselines span both classical and recent deep learning paradigms.
\begin{itemize}
    \item DLinear \cite{zeng2023transformers}: A decomposition-based linear model that separately predicts trend and seasonal components, offering strong performance with minimal complexity.
    \item Transformer \cite{vaswani2017attention}: A self-attention-based model that captures long-range temporal dependencies, widely adapted from NLP tasks.
    \item Autoformer \cite{wu2021autoformer}: Enhances Transformers with series decomposition and auto-correlation mechanisms to better capture seasonality and trend structures.
    \item LightTS \cite{campos2023lightts}: A lightweight Transformer model optimized for efficiency by incorporating grouped attention mechanisms with minimal performance loss.
    \item Reformer \cite{kitaev2020reformer}: Introduces locality-sensitive hashing for sparse attention, enabling memory-efficient modeling of long sequences.
    \item TimesNet \cite{wu2022timesnet}: Combines temporal blocks with frequency-based representations to extract diverse periodicities, achieving strong performance on various benchmarks.
    \item Time-LLM \cite{jin2023time}. Time-LLM is a framework that adapts pre-trained large language models for time series forecasting without requiring additional training. The approach transforms numerical time series data into text through Series-to-Text Reprogramming, enabling LLMs to perform forecasting tasks via in-context learning.
\end{itemize}

\subsection{Parameter Settings}
The proposed model was implemented using PyTorch. For the TIDES framework, the LLM model we adopted is DeepSeek (DeepSeek-R1-Distill-Llama-8B) with 32 layers. The model dimension was set to 16, with 8 attention heads. The patch length and stride for the patch embedding layer were set to 16 and 8, respectively. We used the past one day's data (96 data points) to predict the future one hour's (4-step-ahead) traffic of all regions.
We employed DeepSpeed with Zero-2 optimization to enable efficient distributed training across multiple GPUs. Mixed precision training (bfloat16) was utilized to reduce memory footprint and accelerate computation. The AdamW optimizer was used with a learning rate of 0.001 and OneCycleLR scheduler. All experiments were conducted with a batch size of 16 for training and evaluation on a Sitonholy Cloud Management platform with four A100 GPUs. The model was trained for 100 epochs with early stopping to prevent overfitting. For baseline methods, we adopted the default parameter settings to make the comparison fair.

\begin{figure*}[ht]
\centering
\includegraphics[width=0.95\textwidth]{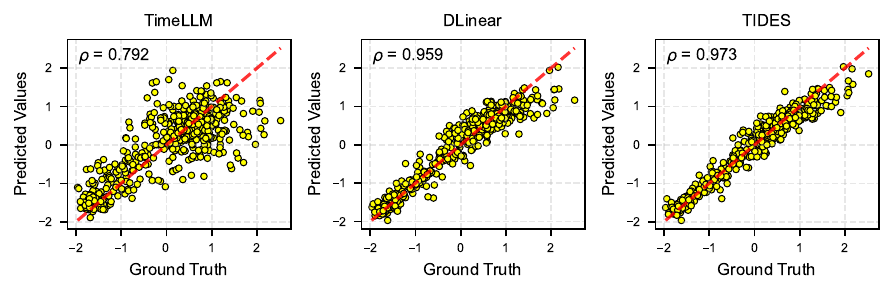}
\caption{Correlation coefficient between predictions and ground truth values. The results from left to right are obtained by Time-LLM, DLinear, and our proposed TIDES, respectively.}
\label{corr_analysis}
\end{figure*}

\subsection{Overall Prediction Performance}

Table \ref{tab:my-table} presents the wireless traffic prediction performance of the proposed TIDES framework against state-of-the-art baseline models explained in the last subsection. The evaluation is conducted across four clusters (A, B, C, and D) using MAE, RMSE, and MAPE. In the table, the best results are highlighted in bold, while the second-best results are underlined.

\begin{figure}[ht]
\centering
\includegraphics[width=0.6\textwidth]{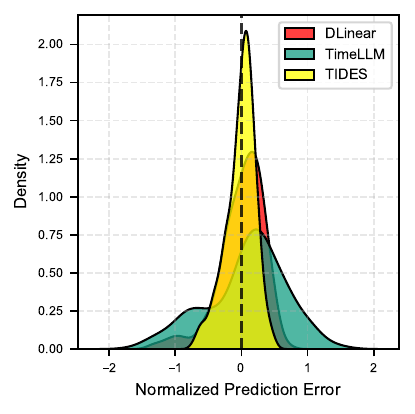}
\caption{Normalized prediction error distribution of different algorithms.}
\label{error_dist}
\end{figure}

Across all zones and evaluation metrics, TIDES consistently outperforms all baselines, achieving the lowest average errors, which indicates its superior prediction accuracy and generalization ability. For example, in Zone A, the proposed TIDES model achieves the best performance across all three evaluation metrics, with MAE (0.2193), RMSE (0.2958), and MAPE (2.7481$\%$), demonstrating strong prediction accuracy and robustness. The second-best performance is consistently obtained by DLinear, which yields an MAE of 0.2330, RMSE of 0.3133, and MAPE of 2.7904$\%$. These results suggest that while DLinear performs competitively, TIDES offers superior capability in capturing the spatiotemporal dynamics of wireless traffic with both lower absolute and relative error.

\begin{figure*}[ht]
        \centering
        \subfloat
        {\label{spatial_compare}
        \includegraphics[width=1.0\textwidth]{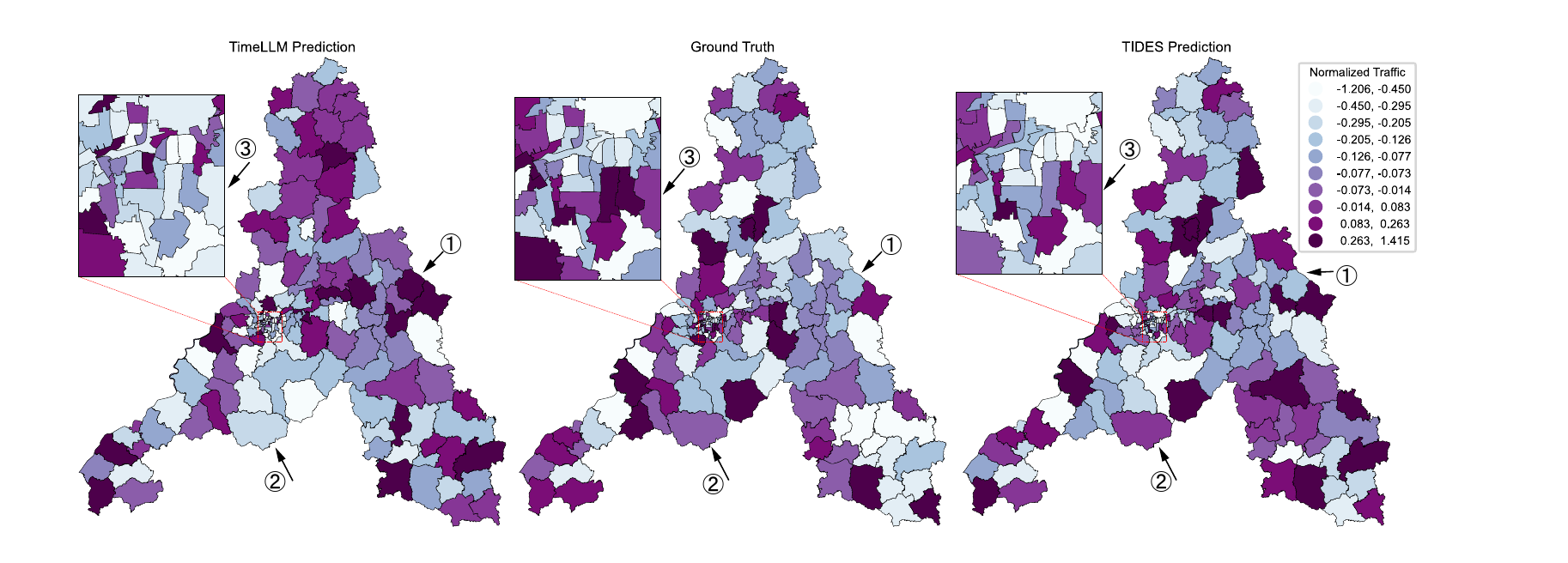}}
        \vfill
        \subfloat{\label{temporal_compare}\includegraphics[width=1.0\textwidth]{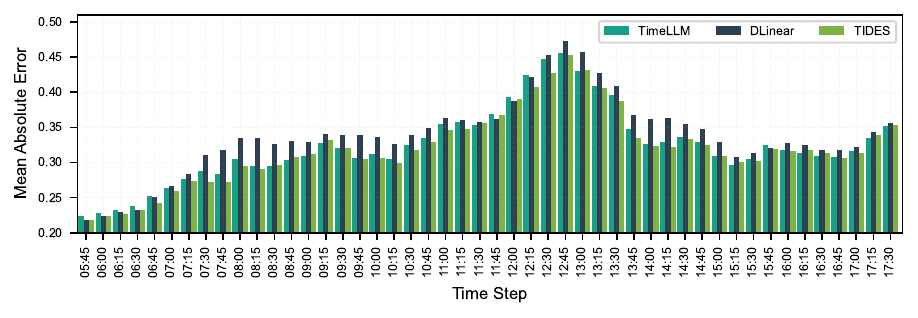}}
        \caption{City-level prediction performance comparison from the spatial view (upper side) and temporal view (lower side).}
        \label{city_level}
\end{figure*}

In Zone B, Time-LLM achieves the best performance in terms of MAE (0.3567) and RMSE (0.5401), demonstrating its effectiveness in absolute and squared error metrics for this particular zone. However, TIDES achieves the lowest MAPE (2.4827$\%$), indicating superior performance in relative error and suggesting it generalizes better when actual traffic volumes vary significantly in magnitude. Furthermore, TIDES ranks second in both MAE (0.4208) and RMSE (0.6459), showcasing a strong balance between absolute and percentage-based prediction accuracy. Autoformer ranks second in MAPE (2.6294$\%$). These results highlight that TIDES provides more stable and scalable accuracy, especially when normalized error (MAPE) is crucial in dynamic traffic environments.

In Zones C and D, TIDES demonstrates strong dominance in wireless traffic forecasting with low error metrics. In Zone C, it achieves the lowest MAE (0.2657), RMSE (0.3621), and MAPE (2.8358$\%$), outperforming TimesNet and DLinear. Similarly, in Zone D, TIDES excels with a low MAE (0.1356), RMSE (0.1746), and MAPE (0.5150$\%$), showcasing its effectiveness in smoother traffic environments. DLinear and TimesNet follow closely in performance. Overall, these results highlight TIDES' ability to deliver precise and stable predictions across varying traffic contexts.

Taken together, these results make a compelling case for TIDES as a versatile and high-performing model. Not only does it achieve the best MAPE across all four zones, it also ranks first in 10 out of the 12 metric-zone combinations. Even in cases where other models like Time-LLM or TimesNet lead in one or two metrics, TIDES remains highly competitive, consistently placing in the top two. This combination of accuracy, robustness, and generalizability positions TIDES as a state-of-the-art solution for city-scale wireless traffic prediction.

\subsection{Correlation Coefficient Performance}
Fig. \ref{corr_analysis} presents a comparative analysis of the prediction accuracy for three different models\footnote{We narrow our analysis to two benchmark methods for clarity, i.e., Time-LLM and DLinear, since these two methods perform relatively well compared with other baselines.}, that is, Time-LLM, DLinear, and the proposed TIDES, through scatter plots of predicted values versus ground truth values. Each subplot includes a red dashed line representing the ideal 1:1 correlation, allowing for visual inspection of model performance. The Pearson correlation coefficient 
$r$ is used as the quantitative metric to assess the linear relationship between the predictions and the ground truth.

The results indicate substantial differences in the correlation performance across the three models. Time-LLM, shown on the left, yields a correlation coefficient of 0.792, suggesting a moderate correlation. The corresponding scatter plot reveals notable dispersion around the identity line, with predictions frequently deviating from the ground truth, indicating limited precision and potential inconsistencies in capturing fine-grained traffic patterns.

In contrast, DLinear significantly improves upon this baseline, achieving a correlation coefficient of 0.959. The points are more tightly clustered around the 1:1 line, suggesting that DLinear provides much more stable and accurate predictions, with reduced error and higher consistency across the dataset.

Most notably, our proposed TIDES model achieves the highest correlation coefficient of 0.973, indicating an extremely strong linear relationship between predicted and actual values. The scatter points align closely with the identity line across the entire value range, demonstrating that TIDES not only captures general trends but also precisely models the magnitude of traffic fluctuations. This result highlights TIDES’ ability to generalize well across varying traffic intensities and affirms its robustness and reliability in urban wireless traffic prediction tasks.

Overall, this correlation analysis validates the superior performance of the TIDES algorithm in aligning with real-world traffic dynamics, outperforming both the transformer-based Time-LLM and the strong baseline DLinear. The high correlation score reinforces the practical utility of TIDES in spatiotemporal traffic modeling and its suitability for deployment in intelligent wireless network management systems.

\subsection{Error Distribution Performance}

Fig. \ref{error_dist} illustrates the normalized prediction error distributions for three models: DLinear, Time-LLM, and our proposed TIDES. The horizontal axis represents the normalized prediction error, while the vertical axis indicates the corresponding probability density. The dashed vertical line at zero denotes the ideal case where the prediction perfectly matches the ground truth.

From the figure, several key observations emerge: 1) TIDES exhibits a sharply peaked and symmetric error distribution centered tightly around zero. This narrow concentration indicates that TIDES consistently produces highly accurate predictions with minimal deviation. The steep peak and thin tails reflect both high precision and robustness, suggesting TIDES makes fewer large errors than its counterparts; 2) DLinear also displays a unimodal distribution centered near zero, but its peak is broader and slightly flatter than that of TIDES. This indicates a slightly higher variance in its prediction errors, implying somewhat reduced accuracy and consistency compared to TIDES, though still outperforming Time-LLM; 3) Time-LLM shows the widest distribution, with a lower and broader peak as well as long tails extending into both negative and positive error ranges. This indicates greater variability and a higher frequency of large prediction errors. The distribution is skewed slightly to the right, suggesting a systematic tendency to overestimate traffic in certain regions.

These results confirm that TIDES achieves the best overall error distribution characteristics, with the smallest variance and highest concentration of prediction errors near zero. Such performance demonstrates TIDES' superior capability in generating accurate and stable predictions, making it highly suitable for practical deployment in real-world wireless traffic forecasting scenarios.

\begin{figure*}[t]
        \centering
        \subfloat
        {\label{cluster0}
        \includegraphics[width=0.95\textwidth]{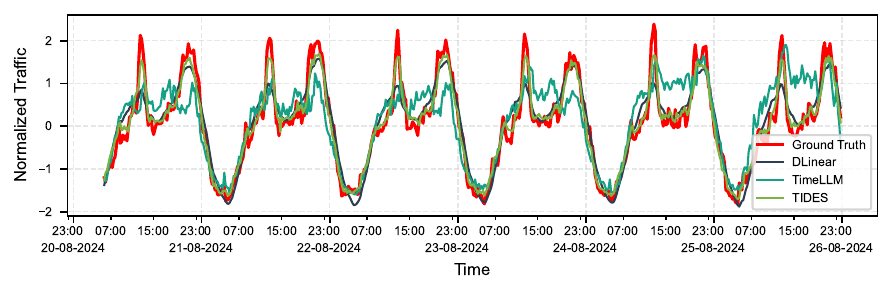}}
        
        \vfill
        \subfloat{\label{cluster3}\includegraphics[width=0.95\textwidth]{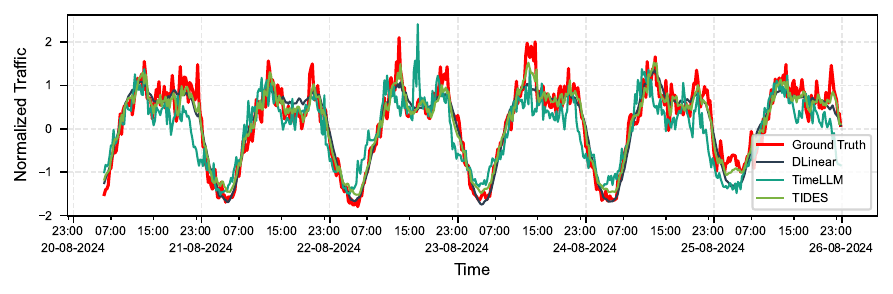}}
        \caption{Prediction versus ground truth along time steps. Upper: Prediction versus ground truth for a randomly selected region of zone A. Lower: Prediction versus ground truth for a randomly selected region of zone C. }
        \label{pred_vs_truth}
\end{figure*}

\subsection{City-level Spatial Temporal Comparisons}
Fig. \ref{city_level} displays the city-level prediction performance comparisons from both spatial and temporal views. Specifically, Fig. \ref{spatial_compare} presents a comprehensive spatial comparison of wireless traffic prediction results from Time-LLM (left) and our proposed TIDES (right), using the ground truth data (center) as a benchmark. Normalized traffic values are categorized into ten levels and represented using a color gradient from light blue (low traffic) to dark purple (high traffic). This visual framework enables intuitive assessment of each model’s spatial prediction performance across the city.

Both models show the capability to capture broad spatial traffic trends; however, TIDES consistently demonstrates stronger spatial alignment with the ground truth. In particular, TIDES better reproduces the spatial distribution of both high- and low-traffic regions, preserving the heterogeneity observed in real-world data. In contrast, Time-LLM tends to overestimate traffic in multiple regions, as indicated by an overrepresentation of darker purple shades.

\begin{figure}[ht]
\centering
\includegraphics[width=0.6\textwidth]{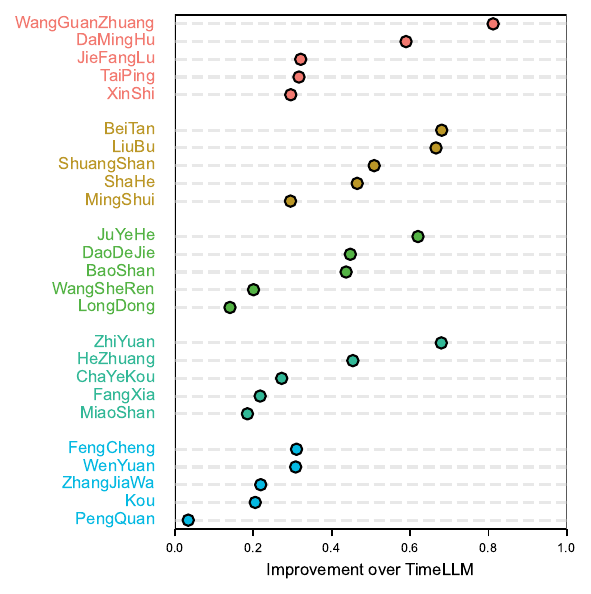}
\caption{Performance improvement over Time-LLM.}
\label{improve_over_base}
\end{figure}

For the first region that is located in the eastern urban edge, the TIDES prediction closely matches the ground truth, accurately capturing moderate-to-high traffic zones. Time-LLM, however, predicts uniformly high traffic, overlooking the nuanced spatial variations. TIDES thus better reflects the spatial heterogeneity typical of urban fringe areas. For the second region, which is located in a suburban area, the ground truth reveals a heterogeneous mix of traffic intensities in this region. TIDES effectively captures this spatial mosaic, including both traffic hotspots and lower-intensity zones. Time-LLM’s predictions are less consistent, often overestimating or underestimating traffic levels and failing to reflect the intricate spatial patterns present in the ground truth. While for the last region which locates in the urban center, the central business district poses a significant challenge due to dense development and rapid traffic transitions. TIDES excels in this high-resolution environment, accurately reflecting sharp spatial transitions and complex local variations. Time-LLM, in contrast, yields a more homogenized prediction that does not adequately differentiate between adjacent high- and low-traffic zones.

The spatial analysis reveals that TIDES achieves superior accuracy across all urban typologies from dense city centers to suburban belts and peripheral zones. Its ability to model both global distribution patterns and local variations marks a significant advancement over existing methods. These strengths highlight the potential of TIDES for high-resolution traffic modeling in practical applications such as urban planning, network optimization, and intelligent transportation systems.

Fig. \ref{temporal_compare} presents a comparative analysis of prediction accuracy among Time-LLM, DLinear, and our proposed TIDES framework, measured by MAE across different time intervals. Notably, TIDES consistently outperforms the baseline models, maintaining the lowest MAE throughout the entire observation period. All three models exhibit increased prediction errors during peak traffic hours (approximately 11:30–13:45), yet TIDES demonstrates superior robustness, particularly under these challenging conditions. This indicates the effectiveness of TIDES in accurately capturing complex temporal dynamics in wireless traffic patterns.

\subsection{Region-Level Prediction Performance}
\subsubsection{Prediction versus ground truth}
Fig. \ref{cluster0} presents a comparative analysis of normalized traffic prediction results over a continuous seven-day window, encompassing both diurnal and weekly variations. This figure contrasts the predicted traffic series generated by three representative models, i.e., DLinear, Time-LLM, and the proposed TIDES algorithm, against the actual ground truth.

The TIDES model exhibits superior alignment with the ground truth across all temporal segments, including sharp transitions during peak hours and low-traffic intervals. In particular, TIDES demonstrates a strong capacity to capture rapid traffic surges and fluctuations, maintaining high temporal resolution and minimal phase lag. In contrast, the DLinear model frequently underestimates traffic during peak periods and exhibits diminished responsiveness to dynamic variations, suggesting limited capability in modeling nonlinear temporal dependencies. Time-LLM, while showing improved adaptability over DLinear, suffers from slight delays in response to abrupt traffic changes and tends to smooth out fine-grained variations, which may lead to attenuated predictions in high-variance periods.

Overall, the visual evidence underscores the robustness of TIDES in modeling complex spatio-temporal dynamics inherent in cellular traffic data. Its improved accuracy can be attributed to its integration of temporal structure awareness and context adaptation mechanisms, which allow it to outperform both conventional linear models and time-aware large language model-based predictors.

\begin{table}[]
\centering
\caption{The generalization ability performance of TIDES over different scenarios.}
\label{tab:generalization}
\renewcommand\arraystretch{1.3}
\resizebox{0.7\columnwidth}{!}{%
\begin{tabular}{@{}c|lll@{}}
\toprule
        Scenarios                     & MAE    & RMSE   & MAPE   \\ \midrule
Zone A $\rightarrow$ Zone A & $0.2255$ & $0.2935$ & $1.6459$ \\
Zone B $\rightarrow$ Zone A & $0.2595(\downarrow15.0\%)$ & $0.3275 (\downarrow 11.5\%)$ & $1.4762(\uparrow 10.3\%)$ \\
Zone C $\rightarrow$ Zone A & $0.2456(\downarrow 8.9\%)$ & $0.3105(\downarrow 5.7\%)$ & $1.6915(\downarrow 2.7\%)$ \\
Zone D $\rightarrow$ Zone A & $0.2328(\downarrow 3.2\%)$ & $0.2996(\downarrow 2.0\%)$ & $1.3675(\uparrow 16.9\%)$ \\ \bottomrule
\end{tabular}%
}
\end{table}

Fig. \ref{cluster3} presents the prediction results on a separate cluster, further validating the generalization capability of the proposed TIDES model. Similar to the previous setting, TIDES maintains the closest alignment with the ground truth across varying traffic intensities and temporal patterns. While both DLinear and Time-LLM demonstrate reasonable performance, they exhibit higher deviation during traffic peaks and valleys. The consistent advantage of TIDES in this scenario reinforces its robustness and adaptability across diverse traffic environments.

\subsubsection{Improvement over baselines}

Fig. \ref{improve_over_base} compares the performance of the proposed TIDES algorithm with Time-LLM across 25 regions. TIDES consistently outperforms Time-LLM, with improvement magnitudes varying by region. The largest gains (up to 0.8) occur in complex, high-traffic areas such as QuanFu and DongGuan, while moderate (0.4–0.6) and smaller ($<$ 0.3) improvements are observed in mixed-density and stable regions, respectively. We can conclude from these results that TIDES shows robust and consistent improvement over Time-LLM, especially in complex or high-traffic regions, making it a more effective solution for spatially diverse traffic prediction tasks.

\subsection{Generalization and Interpretability Analysis}
The generalization ability of TIDES is reported in Table \ref{tab:generalization}, where we train TIDES using data from the source Zone and directly test its performance in the destination Zone without fine-tuning or retraining the model. Noted that here we only sample five regions from each region to accelerate comparisons. We can observe from Table 
\ref{tab:generalization} that performance degradation is small (mostly $<10\%$) across zones, indicating that TIDES generalizes well even under spatial distribution shifts.

\section{Conclusion}
In this paper, we introduced TIDES, a novel traffic prediction framework that integrates large language models with spatial-temporal modeling for improved wireless network management. TIDES addresses the shortcomings of existing approaches by combining regional clustering, prompt-based representation, and cross-domain attention for accurate city-level traffic forecasting. TIDES partitions the network into spatially correlated regions and fine-tunes a shared LLM backbone for each region. By employing prompt engineering, we structure traffic data with statistical features to enhance the model's reasoning capabilities. The DeepSeek module further captures spatial dependencies through attention mechanisms across neighboring regions. Additionally, by freezing core LLM layers and adapting peripheral components, TIDES achieves efficient training and domain adaptation. Our experiments on real-world city-scale datasets show that TIDES outperforms state-of-the-art baselines across multiple metrics, demonstrating that incorporating spatial intelligence into LLM-based predictors enhances accuracy and robustness in dynamic traffic environments. These findings underscore the potential for scalable, prediction-driven resource optimization in future AI-native 6G networks.

In the future, TIDES is expected to evolve into a foundational module for intelligent network control systems. Potential future directions include integrating real-time feedback loops for online adaptation, expanding to multi-modal traffic inputs, and further generalizing the approach to support cross-city or global-scale prediction tasks. Ultimately, the TIDES framework brings us a step closer to fully self-optimizing wireless networks that are capable of anticipating user demand and autonomously optimizing resource provisioning in an increasingly complex communication landscape.

\bibliographystyle{IEEEtran}
\bibliography{llmtp}

\vfill

\end{document}